\documentclass[conference]{IEEEtran}
\IEEEoverridecommandlockouts
\usepackage{cite}
\usepackage{amsmath,amssymb,amsfonts}
\usepackage{algorithmic}
\usepackage{graphicx}
\usepackage{textcomp}
\usepackage{xcolor}
\def\BibTeX{{\rm B\kern-.05em{\sc i\kern-.025em b}\kern-.08em
    T\kern-.1667em\lower.7ex\hbox{E}\kern-.125emX}}

\begin{document}

\title{A Feature-based Approach for the Recognition of Image Quality Degradation in Automotive Applications 
\thanks{Co-funded by \textit{Bayerisches Staatsministerium für Wirtschaft, Landesentwicklung und Energie} (stmwi.bayern.de).}
}


\author{
\IEEEauthorblockN{Florian Bauer}
\IEEEauthorblockA{\textit{b-plus technologies GmbH}\\
Deggendorf, Germany \\
florian.bauer@b-plus.com}
}

\maketitle

\begin{abstract}
Cameras play a crucial role in modern driver assistance systems and are an essential part of the sensor technology for automated driving.
The quality of images captured by in-vehicle cameras highly influences the performance of visual perception systems.
This paper presents a feature-based algorithm to detect certain effects that can degrade image quality in automotive applications.
The algorithm is based on an intelligent selection of significant features. 
Due to the small number of features, the algorithm performs well even with small data sets.
Experiments with different data sets show that the algorithm can detect soiling adhering to camera lenses and classify different types of image degradation.
\end{abstract}


\section{Introduction}
\label{sec:intro}

The functionality of in-vehicle visual perception systems relies on the quality of the images captured by the cameras.
The performance of visual perception algorithms is known to degrade significantly in adverse conditions.
Adverse weather conditions can be rain, snow, or fog. Further adverse conditions can be caused by poor lighting, or simply soil, ice, dust, or water adhering to the camera's lens.

Unlike cameras behind the windshield, fisheye cameras are typically mounted at a low position on a vehicle and are directly exposed to the environment. 
Thus, the lenses of these cameras tend to get soiled, e.g., by splashes of mud or water or by raindrops forming on the lens \cite{Yogamani_2019_ICCV, Uricar2019}. 
Because of their extensive field of view, which typically covers areas with varying lighting conditions, images from fisheye cameras are also susceptible to local overexposure or underexposure. Soiling and incorrect exposure can degrade the quality of images significantly.

Algorithms that can detect image quality degradation are essential to higher levels of vehicle automation. They allow the system to issue warnings about malfunction or even to disable affected functions \cite{Uricar2019, Uricar2019b, RaviKumar2021}.
However, these algorithms may also be applied in the development process of the automation system. 
Poor image quality may cause unexpected, challenging, or dangerous situations, often referred to as corner cases \cite{Bogdoll2021}.
It is essential to detect such situations and include them in the data that is used for the development of machine learning algorithms.

\begin{figure}[htbp]
    \begin{minipage}[b]{.325\linewidth}
        \centering
            \centerline{\includegraphics[width=\linewidth]{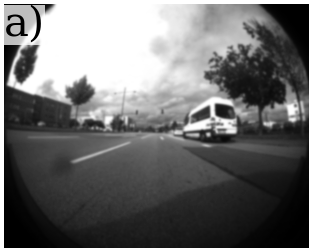}}
    \end{minipage}
    \hfill
    \begin{minipage}[b]{0.325\linewidth}
        \centering
            \centerline{\includegraphics[width=\linewidth]{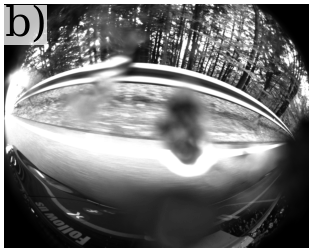}}
    \end{minipage}
    \hfill
    \begin{minipage}[b]{.325\linewidth}
        \centering
            \centerline{\includegraphics[width=\linewidth]{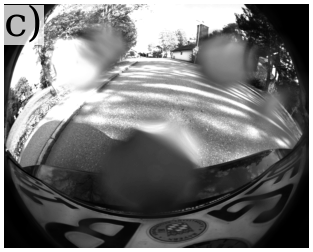}}
    \end{minipage}
    \begin{minipage}[b]{0.325\linewidth}
        \centering
            \centerline{\includegraphics[width=\linewidth]{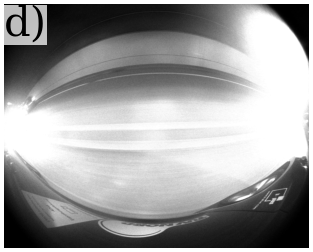}}
    \end{minipage}
    \hfill
    \begin{minipage}[b]{.325\linewidth}
        \centering
            \centerline{\includegraphics[width=\linewidth]{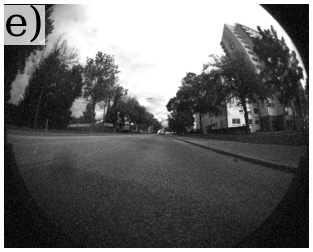}}
    \end{minipage}
    \hfill
    \begin{minipage}[b]{0.325\linewidth}
        \centering
            \centerline{\includegraphics[width=\linewidth]{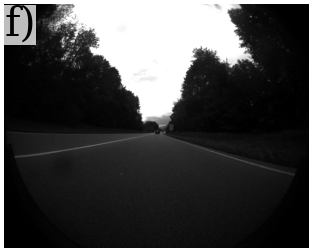}}
    \end{minipage}
    \caption{Sample images from a data set along with typical effects that can degrade image quality: a) blur, b) mud, c) raindrops, d) glare, e) noise, f) underexposure.}
    \label{fig:sample_images}
\end{figure}

Fig.~\ref{fig:sample_images} shows a selection of images (converted to grayscale) captured by four surround-view fisheye cameras on a test vehicle, along with some common effects that can degrade image quality.

The work presented here focuses on the recognition of image quality degradation in automotive applications, caused by soiling adhering to the lens (raindrops, dust, mud, ice), noise, blur, underexposure, or glare.
The key contribution of this paper is the development of a feature-based approach. 
This approach is based on a set of features that can be computed from a single grayscale image.
The features are derived from the statistical moments of the distributions of quantities that result from the application of different filter operations on the grayscale image.

\section{Related Work}
\label{sec:related_work}

The recognition and detection of reduced image quality have been studied in various fields of engineering. 
For instance, in image quality assessment (IQA) the appearance of an image to a human viewer is most important \cite{Mittal2012} and a score may be used to quantify quality. Whereas for images used in visual perception, the ``quality'' refers to the amount of useful information the image contains, e.g., the number of identifiable objects \cite{Tomasi2021}.
Depending on the application, different terminology has been introduced to define image quality and its degradation.  
This section briefly introduces this terminology along with some previously developed methods to evaluate and detect image quality degradation.

In IQA, the algorithms presented in \cite{Mittal2012, Mittal2013} use scene statistics of ``locally normalized luminance coefficients'' to quantify possible losses of ``naturalness'' in the image due to the presence of ``distortions''. 
Here, the term distortion refers to artificial blur, noise, or the result of extensive compression that affects image quality.
Based on the statistics, a representative feature vector is computed, which is used to evaluate the perceptual quality of an image. 

In visual perception applications, adverse weather conditions (e.g., fog, mist, rain, and snow) are known to degrade the performance of many image- and video-based algorithms. 
Often, adverse weather conditions not only reduce the visible range of on-board cameras and cause a loss of contrast, but also cause soiling on the camera's lens. 
In the past, the effect of rain in particular has been studied in this context. 
The detection of raindrops has for instance been studied in \cite{Kurihata2007, Roser2009, Gormer2009, Cord2014, Ishizuka2017}. The survey paper \cite{Hamzeh2021} summarizes methods for the detection of raindrops adhering to a vehicle’s windshield.

However, soiling on lenses of in-vehicle cameras may not only be caused by raindrops. 
In \cite{Einecke2014} a method to detect “image artifacts” that arise from raindrops, dirt, and scratches on the camera lens is discussed. The method is based on a pixel-wise correlation between several frames captured on moving systems.

Recent approaches use \textit{Convolutional Neural Networks} (CNNs) to detect ``soiling'' on camera lenses.
Different networks that can detect ``opaque'' and ``transparent'' soiling on image-, tile-, or pixel-level are presented in \cite{Uricar2019, Das2019, Das2020, RaviKumar2021}.
Although the networks are ``elatively small-sized'' data sets with many thousands of annotated images are used for training (e.g., \emph{WoodScape} \cite{Yogamani_2019_ICCV}). 

Before the rise of deep learning, feature-based approaches were commonly used in computer vision applications.
These approaches typically use the statistics of handcrafted features and a classical machine learning model to perform the final classification \cite{Szeliski2022}. 
E.g., the well-known method in \cite{Dalal2005} proposes features based on the ``histogram of gradients'' (HoG) that enables high accuracy for human detection.
In \cite{Roser2008}, a method is presented that uses features derived from histograms of different image properties (brightness, contrast, sharpness, hue, and saturation). The histograms are computed for multiple regions of interest in the image and the features are compiled into one large vector. A support vector machine (SVM) classifier is applied to estimate the weather situation (clear weather, light rain, heavy rain) based on the vector. 

\section{Feature Extraction}
\label{sec:feautre_extraction}

\begin{figure}[htbp]
    \centerline{\includegraphics[width=\linewidth]{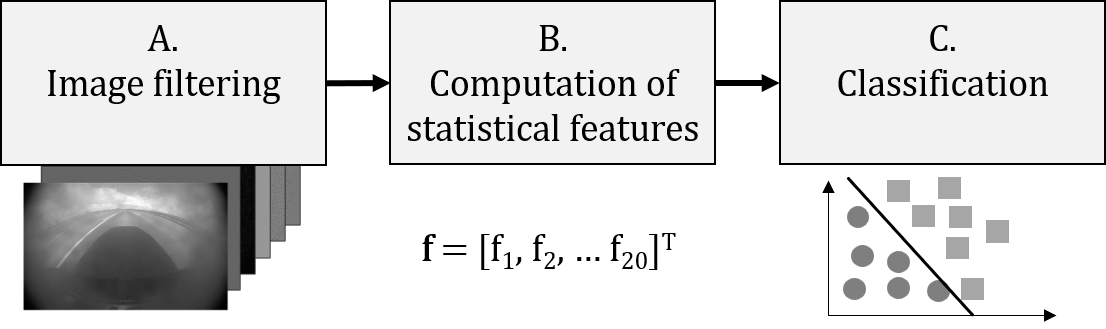}}
    \caption{Overview of the three major steps of the algorithm.}
    \label{fig:process}
\end{figure}

This paper presents a feature-based algorithm that can recognize common effects of image quality degradation in automotive applications. The proposed set of features can be derived from a single grayscale image. 
An SVM is used to recognize different classes of degradation effects based on the features.
Fig.~\ref{fig:process} gives an overview of the algorithm. 
First, different filters are applied to the grayscale image to determine local quantities for blur, sharpness, and others. Next, features based on the statistical distribution of these quantities are computed. 
A machine learning algorithm is applied to recognize different classes of image quality degradation.

Although color images may contain additional valuable information, the proposed algorithm uses grayscale images only. This allows the algorithm to be applied to a wider range of cameras given that some of the color filter arrays used in the automotive industry do not provide color information (e.g. RCCC). 

\subsection{Image filtering}

\begin{figure}[tbp]
    \begin{minipage}[b]{.325\linewidth}
        \centering
            \centerline{\includegraphics[width=\linewidth]{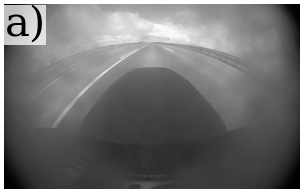}}
    \end{minipage}
    \hfill
    \begin{minipage}[b]{0.325\linewidth}
        \centering
            \centerline{\includegraphics[width=\linewidth]{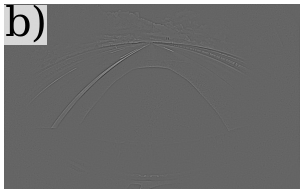}}
    \end{minipage}
    \hfill
    \begin{minipage}[b]{.325\linewidth}
        \centering
            \centerline{\includegraphics[width=\linewidth]{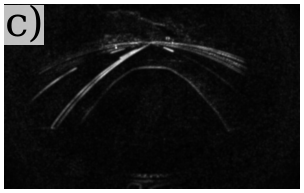}}
    \end{minipage}
    \begin{minipage}[b]{0.325\linewidth}
        \centering
            \centerline{\includegraphics[width=\linewidth]{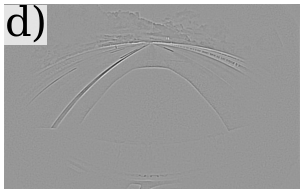}}
    \end{minipage}
    \hfill
    \begin{minipage}[b]{.325\linewidth}
        \centering
            \centerline{\includegraphics[width=\linewidth]{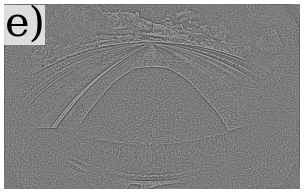}}
    \end{minipage}
    \hfill
    \begin{minipage}[b]{0.325\linewidth}
        \centering
            \centerline{\includegraphics[width=\linewidth]{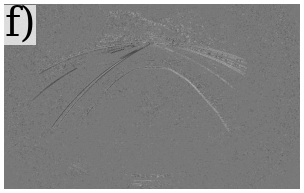}}
    \end{minipage}
    \caption{Initial image and computed quantities: a) grayscale image, b) local mean subtracted field, c) local contrast field, d) gradient field, e) MSCN coefficients, f) products of MSCN coefficients.}
    \label{fig:filtered_images}
\end{figure}

First, a series of filter operations is applied to a grayscale image.
Fig.~\ref{fig:filtered_images} shows a grayscale image captured by a fisheye camera with lens affected by soiling and the five quantities that are computed based on it.
Let $I(i,j)$ with $0 \leq I(i,j) \leq 1$ denote the intensity of each pixel in the initial grayscale image, where $i = 1, 2, \dots H$, and $j = 1, 2, \dots W$ are the pixel indices, and $H$ and $W$ are the height and width of the image, respectively. 

\subsubsection{Local Mean Subtracted Field}

To compute the \textit{local mean subtracted field} ($I(i,j) - \mu(i,j)$) the local mean field $\mu(i,j)$ is required, which is defined by
\begin{equation}
    \label{eq:mu}
    \mu(i,j) = \sum_{k=-K}^K \sum_{l=-L}^L w_{k,l} I(i+k, j+l) \;\; .
\end{equation} 
In this equation, $w$ is a circular Gaussian weighting function sampled out to three standard deviations and scaled to unit volume, as suggested in \cite{Mittal2012, Mittal2013}. $K$ and $L$ determine the size of the filter. For $K = L = 3$ this results in a filter of $7$ pixels by $7$ pixels.
The operation defined by (\ref{eq:mu}) returns the weighted sum of pixel values of the grayscale image within a small neighborhood of a pixel and is commonly used to blur images.
The local mean field $\mu$ is subtracted pixel-wise from the grayscale image to obtain the local mean subtracted field ($I - \mu$).

\subsubsection{Local Contrast Field}

This quantity is defined as:
\begin{equation}
    \label{eq:sigma}
    \sigma (i,j) = (\sum_{k=-K}^K \sum_{l=-L}^L w_{k,l} (I(i+k, j+l) - \mu(i,j))^2)^\frac{1}{2} \; .
\end{equation}
It gives the local variance of an image and can be used as a measure for local image sharpness \cite{Mittal2013}. 

\subsubsection{Gradient field}

The edges in an image are known to be important for visual perception such as object detection but can also be affected by different types of image degradation. Edges in images can be identified by various measures, like the magnitude of the first-order gradients in the horizontal and vertical directions (Sobel), or the Laplacian of an image.
Experiments with different filters show that the Laplacian operator yields the most distinctive features for soiled images.
In Cartesian coordinates, the Laplacian operator gives the sum of the second-order derivatives in horizontal and vertical directions of an image and can be approximated by a discrete filter.

\subsubsection{Mean Subtracted Contrast Normalized (MSCN) Coefficients}

This quantity is inspired by image quality assessment (IQA) algorithms \cite{Mittal2012}.
It is defined as:
\begin{equation}
    \label{eq:mscn}
    \hat{I}(i,j) = \frac{ I(i,j) - \mu(i,j) }{ \sigma(i,j) + e } \;\; , 
\end{equation}
where $e$ is a constant that prevents division by zero.
These coefficients have been observed to follow a certain statistical distribution for images that contain little or no distortion. The presence of image distortion (noise, blur, artifacts from compression) alters the distribution of the coefficients significantly.

\subsubsection{Pair-Wise Product of MSCN coefficients}

Similarly, the products of pairs of MSCN coefficients provide valuable information on image distortion \cite{Mittal2012}. The products $\hat{I}_H(i,j)$ along the horizontal direction can be computed from
\begin{equation}
    \label{eq:mscn_products}
    \hat{I}_H(i,j) = \hat{I}(i,j) \; \hat{I}(i+1,j) \;\; .
\end{equation}

 All the quantities defined in this section can be computed very efficiently and are assumed to provide valuable information to identify image degradation. 
 For instance, the local mean subtracted field ($I - \mu$) provides a local measure of blurriness in the image. In Fig.~\ref{fig:filtered_images} b it can be seen that the mean subtracted field exhibits small magnitudes (color-coded by gray) in the parts of the image that are soiled.
 
\subsection{Computation of Statistical features}

Examining the quantities in Fig.~\ref{fig:filtered_images} showed that their distributions (histograms) exhibit characteristic shapes corresponding to certain effects of image degradation.
Thus, it can be hypothesized that the statistical properties of these distributions can be used to recognize certain types of image quality degradation.

However, the number of features directly influences the complexity of the classification problem. A large number of features may increase the accuracy, but will also increase computation time and the size of the required training set in particular. 
Thus, a small number of highly significant features should be selected. 

The proposed features are the first- and second-order moments (mean and variance) of the distributions of the quantities in Fig.~\ref{fig:filtered_images}. 
They are assumed to be highly significant, because for several distribution functions the shape parameters can be estimated based on these two statistical properties. 
These properties are also often used for functions that do not have a closed form to determine the estimates \cite{Roenko2014, Krupinski2006, Mallat1989}.
Because the distributions may be asymmetric, the moments are calculated for the non-negative and negative values separately.
\begin{equation}
    \label{eq:feature_1}
    \mu_{pos} = \frac{1}{H W} \sum_{i=1}^{H} \sum_{j=1}^{W} x(i, j) \;\; \forall \;\; x(i,j) \geq 0
\end{equation}
\begin{equation}
    \label{eq:feature_2}
    V_{pos} = \frac{1}{H W} \sum_{i=1}^{H} \sum_{j=1}^{W} (x(i, j) - \mu_{pos})^2 \;\; \forall \;\; x(i,j) \geq 0
\end{equation}
Equations~(\ref{eq:feature_1}) and (\ref{eq:feature_2}) give the features $\mu_{pos}$ and $V_{pos}$ based on the non-negative values $x \geq 0$. For features $\mu_{neg}$ and $V_{neg}$ the negative values $x < 0$ are used, resulting in four features per distribution. 
Only two features are computed for the distributions $I(i,j)$ and $\sigma(i,j)$ because they merely have non-negative values.
In total, this results in a feature vector $\mathbf{f}$ with $20$ scalar elements.

\subsection{Classification}

The feature vector $\mathbf{f}$ is used as input for a classifier that maps from feature space into classes. 
For this task, supervised machine learning algorithms such as Decision Trees, Neural Networks, or Support Vector Machines can be used.
Due to its robustness, computational efficiency, and applicability to small and imbalanced data sets, an SVM with an RBF (Radial Basis Function) kernel is used. The algorithm is implemented in \textit{Python} based on the libraries contained in \textit{Scikit-learn} \cite{Pedregosa2011}.

\section{Experiments and Results}
\label{sec:experiments}

\subsection{Data Sets}

Table~\ref{tab:datasets} provides information on each of the four data sets used for the experiments in this section.
\begin{table}
    \caption{Overview of data sets used in the experiments.}
    \begin{center}
        \begin{tabular}{llllll}
        \hline
        \multicolumn{2}{l}{\textbf{Data set }}     & \textbf{1} & \textbf{2} & \textbf{3} & \textbf{4}  \\
        \multicolumn{2}{l}{\textbf{Resolution }}   & 533x400    & 640x480   & 1280x960   & 533x400           \\
        \multicolumn{2}{l}{\textbf{\# Classes }}   & 6          & 2          & 2          & 2           \\
        \multicolumn{2}{l}{\textbf{\# Images }}    & 1606       & 3152       & 13234      & 1616            \\ 
         & Clean                                   & 513        & 1802       & 8234       & 843         \\
         & Soiled                                  & 196        & 1350       & 5000       & 773         \\
         & Blur                                    & 256        & -          & -          & -           \\
         & Glare                                   & 113        & -          & -          & -           \\
         & Noise                                   & 256        & -          & -          & -               \\
         & Underexp.                               & 272        & -          & -          & -               \\
        \hline
        \end{tabular}
    \end{center}
    \label{tab:datasets}
\end{table} 
 All data sets contain at least images from two classes: \textit{clean} and \textit{soiled}. Data Set 1 has additional images according to four more effects of image degradation.
 The images in Data Set 1 were sampled from video recordings captured by four surround-view fisheye cameras which are mounted on the front and rear of a test vehicle as well as on both mirrors. Sample images can be seen in Fig.~\ref{fig:sample_images}.
 The images in Data Set 2 were extracted from a collection of videos that were taken during every day driving situations by a rear view camera in a modern production car. The images clearly exhibit some artifacts originating from the lossy compression that was applied to the video data.
 Data Set 3 is the subset of the WoodScape data set that is publicly available \cite{Yogamani_2019_ICCV}.
 Data set 4 is a combination of data sets 1, 2, and 3. The images are selected in a way that the three data sets and two classes are roughly balanced.

\subsection{Recognition of soiled images}

The proposed algorithm is applied to the four data sets to classify the input images as either \textit{clean} or \textit{soiled} (binary classification). 
The data set is split into training ($75\%$) and test data ($25\%$).
The features are mean removed and scaled to unit variance first.
An SVM classifier with an RBF kernel is trained on each of the four data sets. 
The regularization parameter $C$ and the kernel coefficient $\gamma$ (see \cite{Chang2011}) are adjusted for each of the data sets individually. 
 
Table~\ref{tab:accuracy} summarizes the performed experiments and accuracies achieved on the test data.
\begin{table}[htbp]
    \caption{Summary of experiments performed on different data sets.}
    \begin{center}
        \begin{tabular}{lllll}
            \hline                                                      
            \textbf{Data set}           & \textbf{1}   & \textbf{2}   & \textbf{3}   & \textbf{4}   \\
            \textbf{Train size}         & 531   & 2364  & 9925  & 1212  \\
            \textbf{Test size}          & 178   & 788   & 3309  & 404   \\
            \textbf{Accuracy [$\%$]}    & 95.51 & 99.75 & 98.70 & 96.53 \\
            \hline
        \end{tabular}
    \end{center}
    \label{tab:accuracy}
\end{table}
The accuracy is defined as the ratio of the number of correctly predicted test images to the total number of test images.

One of the major advantages of the proposed, feature-based algorithm is its low complexity, because the classification of the images is based on merely $20$ features. 
Even for training sets with only $500$ images, accuracies $\geq 95\%$ can be achieved. 

\subsection{Accuracy Compared to CNN}

To compare the feature-based approach to a CNN (see \cite{Uricar2019}), the same experiments as described in \cite{Uricar2019} are performed. 
In these experiments, different subsets of the WoodScape data set are used: (I) training and testing on images captured by front (FV) and rear (RV) cameras only, (II) training on FV and RV cameras and testing on all cameras, and (III) training and testing on all cameras. The results are shown in Table~\ref{tab:uricar}.
\begin{table}[htbp]
    \caption{Summary and comparison of experiments performed with the feature-based approach and a CNN on different subsets of data set 3.}
    \begin{center}
        \begin{tabular}{llll}
        \hline                                                      
        \textbf{Experiment}             & (I)       & (II)       & (III)$^{\mathrm{a}}$    \\
        \textbf{Training data}          & FV and RV & FV and RV & all   \\
        \textbf{Test data}              & FV and RV & all       & all   \\
        \textbf{Accuracy$^{\mathrm{b}}$ {[}\%{]}}      & 99.39     & 91.76     & 98.70 \\
        \textbf{Accuracy CNN {[}\%{]}}  & 98.83     & 83.97     & 99.93 \\
        \hline
        \multicolumn{4}{l}{$^{\mathrm{a}}$Same experiment as shown in Table~\ref{tab:accuracy}, data set 3.} \\
        \multicolumn{4}{l}{$^{\mathrm{b}}$Feature-based approach.}
        \end{tabular}
    \end{center}
    \label{tab:uricar}
\end{table}
The accuracy achieved by the CNN is given in the bottom row. It is computed based on the raw confusion matrices presented in \cite{Uricar2019}.
The results of experiment (II) provide insight into how well the algorithm generalizes across images taken by cameras mounted on different positions (different perspectives, e.g. Fig.~\ref{fig:sample_images}, a and b).
Although the number of images available for the training of the algorithm is significantly smaller, the feature-based approach outperforms the CNN in this task.

\subsection{Classification of Degradation Effects}

\begin{figure}[htbp]
    \centerline{\includegraphics[width=\linewidth]{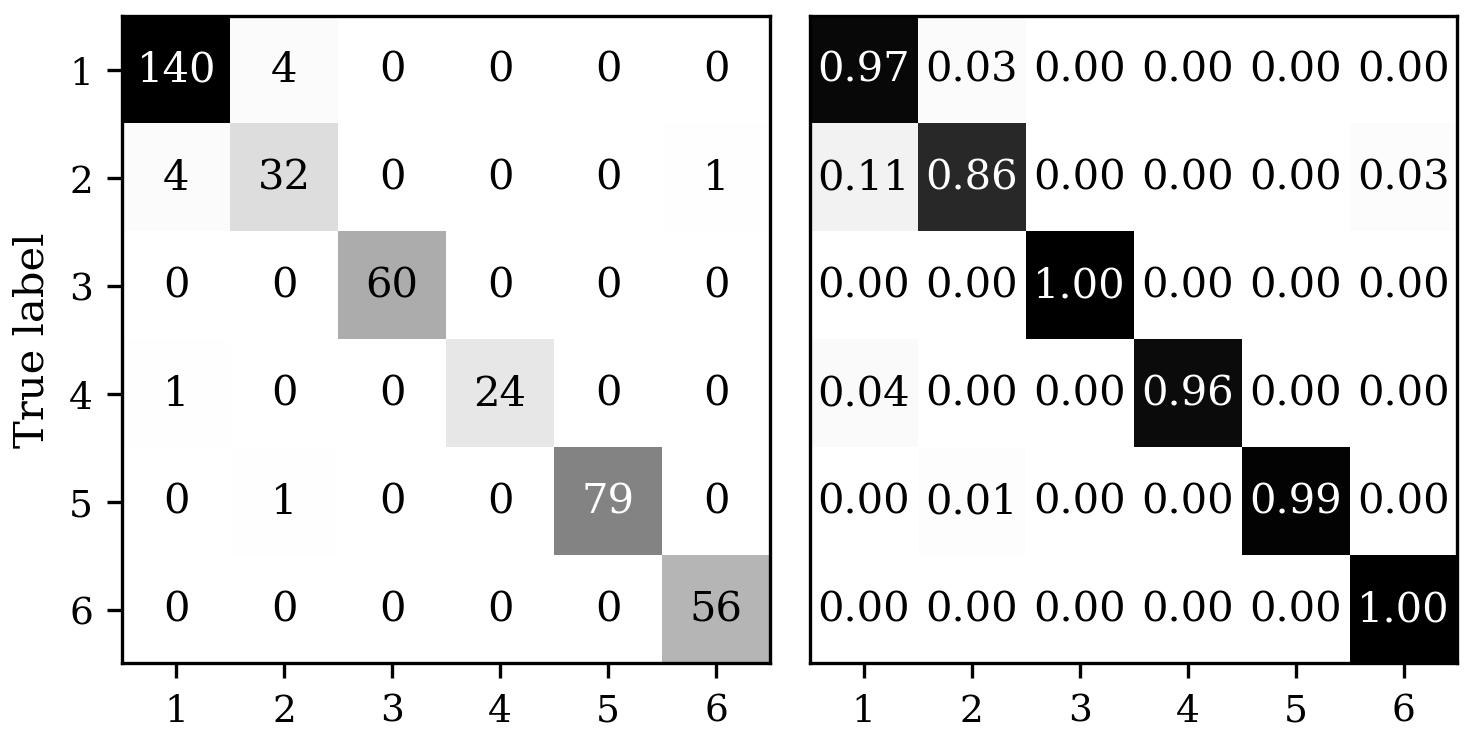}}
    \caption{Raw (left) and normalized (right) confusion matrices (normalized over the true conditions, i.e. rows). 1: clean, 2: soiled, 3: blur, 4: glare, 5: noise, 6: underexposure.}
    \label{fig:conf_matrix_0}
\end{figure}

In this experiment the proposed algorithm is applied to the entire Data Set 1 (multi-class classification). 
The results presented in Fig.~\ref{fig:conf_matrix_0} show that the feature-based algorithm is able to recognize more than just soiled images.
 Other degradation effects can be classified even more accurately. The algorithm achieves an overall accuracy of $96.52\%$.

\section{Conclusion}
\label{sec:conclusion}

This paper presents a feature-based algorithm that is based on an intelligent selection of features that are significant for the detection of image quality degradation. 
From experiments with different image data sets it is found that the algorithm can reliably detect soiling on camera lenses. 
A comparison with another approach (based on Deep Learning) shows that it generalizes better across images captured from different camera positions.
The algorithm is also capable of classifying different effects of image quality degradation, such as soiling, blur, glare, noise, or underexposure.
The major advantages of the presented algorithm are that small data sets suffice to train the classifier and that it can be computed very efficiently.

\newpage

\bibliographystyle{IEEEtran}
\bibliography{IEEEabrv,INSTATE}

\end{document}